\documentclass{article}
\usepackage{times}
\usepackage{latexsym}
\usepackage{tabularx}
\usepackage{multirow}
\usepackage{graphicx}
\usepackage{color}
\usepackage{xcolor}
\usepackage{soul}
\usepackage{balance}
\usepackage{makecell}
\usepackage[utf8]{inputenc}
\usepackage{url}

\author{Felipe Soares \\
Universidade Federal do Rio Grande do Sul\\
Barcelona Supercomputing Center \\
 {\tt fs@felipesoares.net} \\
Martin Krallinger \\
Barcelona Supercomputing Center \\
  {\tt martin.krallinger@bsc.es} \\}

\begin{document}
\title{ BVS Corpus: A Multilingual Parallel Corpus of Biomedical Scientific Texts\\}
\maketitle
\begin{abstract}
The BVS database (Health Virtual Library) is a centralized source of biomedical information for Latin America and Carib, created in 1998 and coordinated by BIREME (\textit{Biblioteca Regional de Medicina}) in agreement with the Pan American Health Organization (OPAS). Abstracts are available in English, Spanish, and Portuguese, with a subset in more than one language, thus being a possible source of parallel corpora. In this article, we present the development of parallel corpora from BVS in three languages: English, Portuguese, and Spanish. Sentences were automatically aligned using the Hunalign algorithm for EN/ES and EN/PT language pairs, and for a subset of trilingual articles also. We demonstrate the capabilities of our corpus by training a Neural Machine Translation (OpenNMT) system for each language pair, which outperformed related works on scientific biomedical articles. Sentence alignment was also manually evaluated, presenting an average 96\% of correctly aligned sentences across all languages. Our parallel corpus is freely available, with complementary information regarding article metadata.

\end{abstract}

\section{Introduction}
The availability of cross-language parallel corpora is one of the basis of current Statistical and Neural Machine Translation systems (SMT and NMT). Acquiring a high-quality parallel corpus that is large enough to train MT systems, specially NMT ones, is not a trivial task, since it usually demands human curating and correct alignment. In light of that, the automated creation of parallel corpora from freely available resources is extremely important in Natural Language Processing (NLP), enabling the development of accurate MT solutions. Many parallel corpora are already available, some with bilingual alignment, while others are multilingually aligned, with 3 or more languages, such as Europarl \cite{koehn2005europarl}, from the European Parliament, JRC-Acquis \cite{steinberger2006jrc}, from the European Commission, OpenSubtitles \cite{zhang2014dual}, from movies subtitles. \par

The extraction of parallel sentences from scientific writing can be a valuable language resource for MT and other NLP tasks. The development of parallel corpora from scientific texts has been researched by several authors, aiming at translation of biomedical articles \cite{wu2011statistical,NEVES16.800}, or named entity recognition of biomedical concepts \cite{kors2015multilingual}. Regarding Portuguese/English and English/Spanish language pairs, the FAPESP corpus \cite{aziz:2011:newfapesp}, from the Brazilian magazine \textit{revista pesquisa FAPESP}, contains more than 150,000 aligned sentences per language pair, constituting an important language resource.\par

In Latin America and Carib, the Pan American Health Organization (OPAS), in agreement with BIREME (\textit{Biblioteca Regional de Medicina)}, maintains the BVS database, which is an important source of biomedical texts in three main languages: English, Spanish, and Portuguese. Currently, BVS has more than 1 million texts indexed, and also provides integrated search capabilities with PUBMED. \par

In this article, we explore the BVS database as a source of parallel corpora for the 3 aforementioned languages. We developed a trilingual parallel corpus with the 3 languages, as well as parallel corpora of English/Portuguese and English/Spanish abstracts. In addition, we provided various metadata regarding the publications.

\section{Licensing}

Most articles in the BVS database are open access documents. In order to avoid any copyright issues, we included in our datasets only open access documents. To retrieve license information, we crawled the BVS website containing information about the indexed journals \footnote{\url{http://portal.revistas.bvs.br/}} as well as the Directory of Open Access Journals \footnote{\url{https://doaj.org/}}.

\section{Material and Methods}
\begin{figure}[h]
\begin{center}
\includegraphics[scale=0.95]{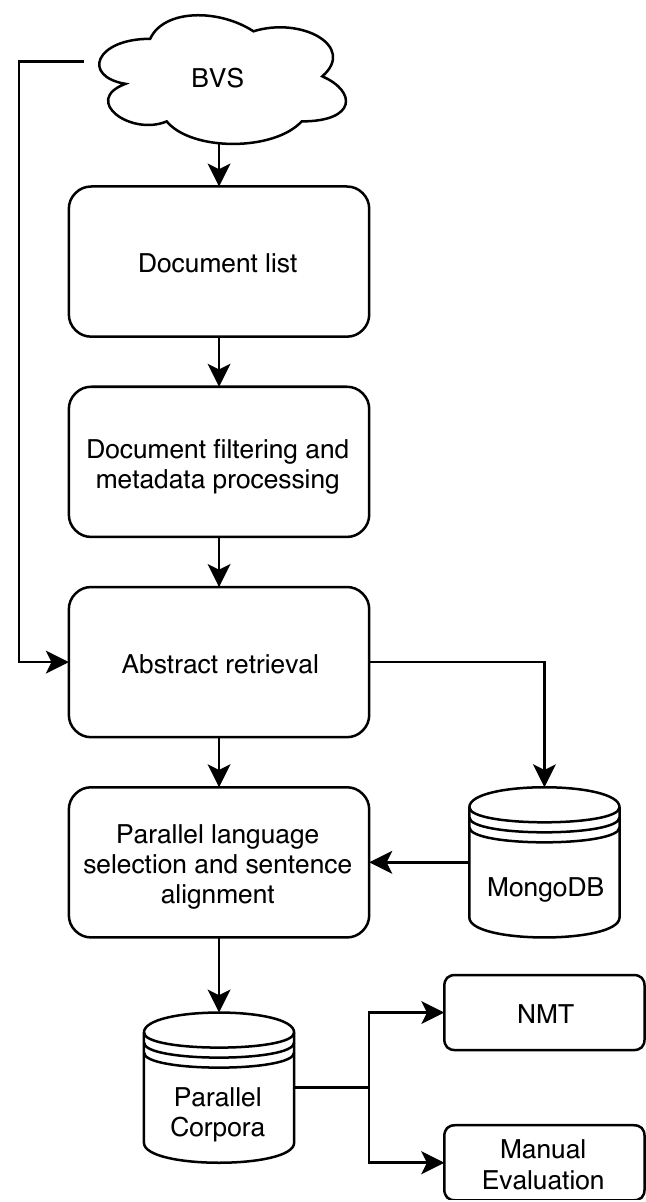} 
\caption{Method employed for corpora building.}
\label{fig_method}
\end{center}
\end{figure}

In this section, we detail the information retrieved from BVS website, the filtering process, the sentence alignment, and the evaluation experiments. Figure \ref{fig_method} shows the diagram of the steps followed for the development of the parallel corpora.

\subsection{Document retrieval and parsing}

BVS's website\footnote{\url{http://bvsalud.org/}} offers simple and advanced search capabilities. We iteratively queried the database to retrieve all lists of results, which were then parsed and all relevant contents stored, such as authorship, title, and abstracts. We adopted  the MongoDB  database system, as it is  document-oriented, and allows for the easy querying and storage of this type of data. \par

After the initial filtering, the resulting documents were processed for language checking\footnote{\url{https://github.com/Mimino666/langdetect}} to make sure that there was no misplacing of abstract language (e.g. English abstracts in the Portuguese field, or the other way around), removing the documents that presented such inconsistency. In addition, we also removed newline/carriage return characters (i.e \textbackslash n and \textbackslash r), as they would interfere with the sentence alignment tool.

\subsection{Sentence alignment}

For sentence alignment, we used the LF aligner tool\footnote{\url{https://sourceforge.net/projects/aligner/}}, a wrapper around the Hunalign algorithm \cite{vargaparallel}, which provides an easy to use and complete solution for sentence alignment, including pre-loaded dictionaries for several languages. \par

Hunalign uses Gale-Church sentence-length information to first automatically build a dictionary based on this alignment. Once the dictionary is built, the algorithm realigns the input text in a second iteration, this time combining sentence-length information with the dictionary. When a dictionary is supplied to the algorithm, the first step is skipped. A drawback of Hunalign is that it is not designed to handle large corpora (above 10 thousand sentences), causing large memory consumption. In these cases, the algorithm cuts the large corpus in smaller manageable chunks, which may affect dictionary building.

The parallel abstracts were supplied to the aligner, which performed sentence segmentation followed by sentence alignment. After sentence alignment, the following post-processing steps were performed: (i) removal of all non-aligned sentences; (ii) removal of all sentences with fewer than three characters, since they are likely to be noise.

\subsection{Machine translation evaluation}
To evaluate the usefulness of our corpus for MT purposes, we trained an NMT model using the OpenNMT system \cite{2017opennmt} for all language pairs. The produced translations were evaluated according to the BLEU score \cite{papineni2002bleu}.

\subsection{Manual evaluation}
\label{sub_manual_eval}
Although the Hunalign algorithm usually presents a good alignment between sentences, we also conducted a manual validation to evaluate the quality of the aligned sentences. We randomly selected 300 sentences, 100 for the triligual subset, and 100 for each subset of EN/PT and EN/ES.  If the pair was fully aligned, we marked it as "correct"; if the pair was incompletely aligned, due to segmentation errors, for instance, we marked it as "partial"; otherwise, when the pair was incorrectly aligned, we marked it as "no alignment".

\section{Results and Discussion}
In this section, we present the corpus' statistics and quality evaluation regarding NMT system, as well as the manual evaluation of sentence alignment.

\subsection{Corpus statistics}

Table \ref{table_stats} shows the statistics (i.e. number of sentences) for the aligned corpus according to the 2 language pairs and the trilingual subset. The dataset is available\footnote{\url{url_will_be_supplied_later}} in TMX format \cite{Rawat2016}, since it is the standard format for translation memories. We also made available the aligned corpus in an SQLite database in order to facilitate future subset selection. In this database, we included the following metadata information: year, keywords in the available languages, database of origin, country, authorship, and URL for the full-text when available. \par

\begin{table}[h]

\begin{center}
\begin{tabular}{|l|r|}
\hline  \multicolumn{1}{|c}{\bf Language Pairs} & \multicolumn{1}{|c|}{\bf Sentences}\\ \hline
EN/PT          & 711,475   \\
EN/ES               & 789,547   \\
EN/PT/ES  & 203,719   \\
\hline
\end{tabular}
\end{center}
\caption{\label{table_stats} Corpus statistics according to language pair. }

\end{table}

\subsection{Translation experiments}
Prior to MT experiments, sentences were randomly split in three disjoint datasets: training, development, and test. Approximately 14,000 sentences were allocated in the development and test sets, while the remaining was used for training. For the NMT experiment, we used the Torch implementation\footnote{\url{http://opennmt.net/OpenNMT/}} to train a 2-layer LSTM model with 500 hidden units in both encoder and decoder, with 20 epochs. During translation, the option to replace UNK words by the word in the input language was used. \par

Table \ref{table_BLEU} presents the BLEU scores for both translation directions with the 3 language pairs for the development and test partitions. We also included the best scores from a similar parallel corpus from Scielo \cite{NEVES16.800} as a benchmark. \par

\begin{table}[h]
\centering

\begin{tabular}{|c|c|c|c|c|}
\hline
\multicolumn{2}{|c|}{\bf Language Pairs} & \makecell{\bf Dev}  		& \makecell{\bf Test} & \makecell{\bf Bench}\\ \hline
\multirow{2}{*}{EN-ES}  & EN$\rightarrow$ES  & 34.80		& 34.96 & 32.75     \\ \cline{2-5} 
                     	& ES$\rightarrow$EN  & 33.82 		& 34.28    & 30.53 \\ \hline
\multirow{2}{*}{PT-ES}  & PT$\rightarrow$ES  & 55.78 			&  56.11    & - \\ \cline{2-5} 
                     	& ES$\rightarrow$PT  & 56.26   			&  56.50 & - \\ \hline
\multirow{2}{*}{EN-PT}  & EN$\rightarrow$PT  & 35.62  		& 36.03   & 33.37 \\ \cline{2-5} 
                     	& PT$\rightarrow$EN  & 35.88   		& 36.12  & 31.78 \\ \hline
\end{tabular}

\caption{\label{table_BLEU} BLEU scores for translation using OpenNMT for the development and test partitions. Previous related work by Neves et al.(2016) is also presented for comparison in the right-hand column as benchmarking.}

\end{table}

Our models achieved better performance than the benchmark for all language pairs and directions, with at least 2.21 percentage points (pp) higher for the EN/ES language pair, achieving a maximum of 4.34 pp for the EN/PT language pair. It is noticeable the high scores achieved in the ES/PT pair, which we expect to be due to the high similarity between both languages.

Below, we demonstrate some sentences translated by OpenNMT compared to the suggested human translation. One can notice that NMT present reasonable results, especially for the PT/ES language pair.

\smallskip

\begin{quote}
Source: \textit{El concepto de necesidades es fundamental para el trabajo de Enfermería.}
\end{quote}

\begin{quote}
Human translation: \textit{The concept of needs is central to the work of Nursing.}
\end{quote}

\begin{quote}
OpenNMT: \textit{The concept of needs is fundamental for nursing work.}
\end{quote}

\smallskip

\begin{quote}
Source: \textit{Não houve associação significante entre o uso de método contraceptivo e as variáveis demográficas e socioeconômicas analisadas.}
\end{quote}

\begin{quote}
Human translation: \textit{No significant association was found between the use of any contraceptive method and demographic and socioeconomic variables.}
\end{quote}

\begin{quote}
OpenNMT: \textit{There was no significant association between contraceptive use and demographic and socioeconomic variables.}
\end{quote}

\smallskip

\begin{quote}
Source: \textit{Método: estudio cualitativo del tipo descriptivo, apoyado en el referencial de Hersey y Blanchard.}
\end{quote}

\begin{quote}
Human translation: \textit{Método: Pesquisa qualitativa do tipo descritiva, apoiada no referencial de Hersey e Blanchard.}
\end{quote}

\begin{quote}
OpenNMT: \textit{Método: estudo qualitativo do tipo descritivo, apoiado no referencial de Hersey e Blanchard.}
\end{quote}

\smallskip

\subsection{Sentence alignment quality}

\begin{figure}[h]
\begin{center}
\includegraphics[scale=0.3]{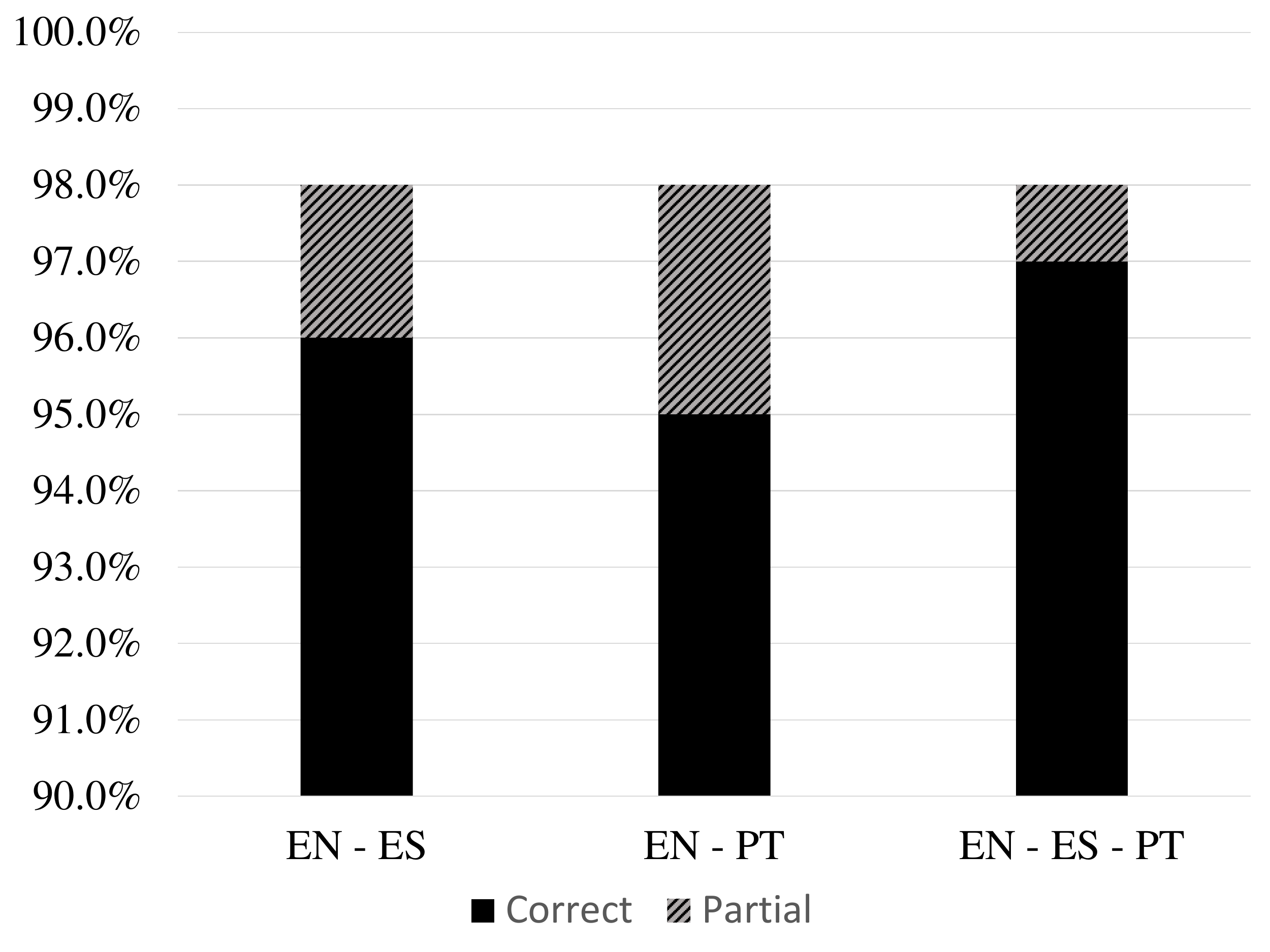} 
\caption{Alignment accuracy for the three language subsets.}
\label{fig_alignment}
\end{center}
\end{figure}

We manually validated the alignment quality for 300 sentences randomly selected from the parsed corpus and assigned quality labels according Section \ref{sub_manual_eval}. From all the evaluated sentences, average 96\% were correctly aligned, while average 2\% were partially aligned. The trilingual subset was the one with the best alignment, achieving 97\% correct alignment. The small percentage of no alignment is probably due to the use of Hunalign algorithm with the provided dictionaries. Figure \ref{fig_alignment} shows the alignment accuracy for all language subsets.\par

\section{Conclusion and future work}

We developed a parallel corpus of biomedical abstracts in English, Spanish, and Portuguese. Our corpus is based on the BVS database, which contains biomedical texts from several sources in Latin America and Carib. The corpus contains the EN/ES, EN/PT language pairs as well as a trilingual subset of EN/PT/ES sentences. \par

Our corpora were evaluated through NMT experiments with OpenNMT system, presenting superior performance regarding BLEU score than a related work with a similar corpus. The NMT model presented remarkable results for the PT/ES language pair, possibly due to the similarity between the languages. We also manually evaluated sentences regarding alignment quality, with average 96\% of sentences correctly aligned. \par

For future work, we foresee the use of the presented corpus in mono and cross-language text mining tasks, such as text classification and clustering. As we included several metadata, these tasks can be facilitated. Other machine translation approaches can also be tested, including the concatenation of this corpus with other multi-domain ones.

\balance
\bibliographystyle{unsrt}
\bibliography{emnlp2018}

\end{document}